\pdfoutput=1
\documentclass[10pt,twocolumn,letterpaper]{article}

\usepackage{cvpr}              
\usepackage{comment}
\usepackage[dvipsnames]{xcolor}

\definecolor{cvprblue}{rgb}{0.21,0.49,0.74}
\usepackage[pagebackref,breaklinks,colorlinks,citecolor=cvprblue]{hyperref}

\def\confName{CVPR}

\title{Solution for 8th Competition on Affective \& Behavior Analysis in-the-wild}

\author{Jun Yu, Yunxiang Zhang, Xilong Lu, Yang Zheng, Yongqi Wang ,Lingsi Zhu\\
University of Science and Technolog of China\\
{{\tt\small harryjun@ustc.edu.cn,} {\tt\small \{mesa,luxilong,zhengyang,wangyongqi,ls-zhu24\}@mail.ustc.edu.cn}}
}

\begin{document}
\maketitle
\begin{abstract}
In this report, we present our solution for the Action Unit (AU) Detection Challenge, in 8th Competition on Affective Behavior Analysis in-the-wild. 
In order to achieve robust and accurate classification of facial action unit in the wild environment, 
we introduce an innovative method that leverages audio-visual multimodal data. Our method employs ConvNeXt as the image encoder and uses Whisper to extract Mel spectrogram features. For these features, we utilize a Transformer encoder-based feature fusion module to integrate the affective information embedded in audio and image features. This ensures the provision of rich high-dimensional feature representations for the subsequent multilayer perceptron (MLP) trained on the Aff-Wild2 dataset, enhancing the accuracy  of AU detection.

\end{abstract}

\section{Introduction}
In recent years, as digital humans and human-computer interaction technologies have become focal points in the field of artificial intelligence\cite{demirel2022digital}, facial expression recognition has garnered increasing attention. These technologies hold the promise of enabling a more accurate understanding of human emotions, thereby facilitating the design of more lifelike digital humans and more user-friendly interaction systems. Derived from the Facial Action Coding System (FACS)\cite{prince2015facial}, Action Unit (AU) are regarded as the fundamental building blocks of facial expressions and play a crucial role in emotion recognition.
The 8th Affective Behavior Analysis in-the-wild (ABAW8)\cite{kollias2022abaw,kollias2023abaw,kollias2023abaw2,Kollias2025,kollias20246th,kollias20247th} Competition aims to address challenges related to human affective behavior analysis. The competition has built a multimodal facial expression dataset, Aff-Wild2\cite{kollias2019deep,kollias2019expression,kollias2020analysing,kollias2019face,kollias2021affect,kollias2021analysing,kollias2021distribution,kollias2024distribution,kolliasadvancements,zafeiriou2017aff}, and introduced the Action Unit (AU) Detection Challenge as one of its tracks. In this challenge, participants are required to detect 12 types of Action Units in every frame of the provided videos.

The Action Unit (AU) recognition challenge faces numerous difficulties\cite{walecki2017deep}. Firstly, the complexity of multimodal data significantly increases the difficulty of the detection task. Second, the uncertain lighting conditions and noise inherent in in-the-wild environments further exacerbate the challenge, placing even greater demands on the robustness of detection methods. Furthermore, the changes in facial actions in videos constitute a dynamic process that requires modeling and analysis over the temporal dimension\cite{8100199}, rather than relying solely on independent processing of static frames.

To address these issues, this study designs and optimizes solutions from the perspectives of feature extraction and enhancement, as well as multimodal fusion. These efforts have led to improved recognition performance and achieved state-of-the-art excellent results. Our contributions are summarized below:

\begin{itemize}
\item  We apply pre-trained Whisper and ConvNeXt to the audio and visual streams respectively, performing end-to-end high-dimensional feature extraction. Simultaneously, we create both global and local views for the video and audio. The local views, obtained by extracting enhanced spatio-temporal crops from the video and enhanced time-frequency crops from the audio, effectively capture fine-grained key features.
\item  We integrate an advanced cross-modal alignment and fusion module that synchronizes audio and visual features. Using multi-scale self-attention and adaptive sliding windows, it captures temporal context and fuses complementary information, thereby enhancing the accuracy and robustness of AU recognition.

\end{itemize}

\section{Method}

In this section,we introduce each component that we have attempted in our approach. Our complete system integrates robust audio-visual feature extraction, advanced cross-modal alignment, and  classification and auxiliary optimization. The pipeline consists of three key modules, described in detail below.

\subsection{Data Preprocessing \&  Feature Extraction}

To extract rich feature representations, we generate both "global" and "local" views from the audio input\cite{sarkar2024xkd}. The global view is constructed using the full spectrogram, which captures broad temporal and spectral information. Specifically, we use the entire log-Mel spectrogram\cite{meng2019speech} as input for the global view, preserving the full context of the audio.

For the local view, we focus on extracting enhanced time-frequency crops from the spectrogram. These crops are small, overlapping sections of the Mel spectrogram that capture fine-grained time-frequency patterns. We randomly sample a segment of the spectrogram  and apply frequency masking as augmentations. This helps capture localized variations in the audio signal, such as specific speech or sound event patterns, which are crucial for distinguishing finer details in the audio stream.

In the audio preprocessing stage, we adopt the efficient approach validated in the Whisper model \cite{radford2023robust}to convert raw audio signals into high-dimensional feature representations. 
All raw audio is first resampled to 16,000\,Hz and globally normalized, mapping the signal amplitude to the range $[-1,1]$in order to minimizes the effects of varying recording conditions.
The normalized audio signal $x(t)$ is segmented using a 25-millisecond window with a 10-millisecond stride to perform the Short-Time Fourier Transform (STFT), yielding the spectral representation $S(t,k)$:
    \[
    S(t,k) = \sum_{n=0}^{N-1} x(n) \cdot w(n-t) \cdot e^{-j \frac{2\pi k n}{N}},
    \]
    where $w(n)$ denotes the window function and $N$ is the window length. An 80-channel Mel filter bank $\mathbf{M} \in \mathbb{R}^{80 \times N}$ is then applied to project the spectrum onto the Mel frequency scale. The log-energy of the projected spectrum is computed as:
    \[
    X(t,f) = \log \left( \sum_{k} M_{f,k} \left| S(t,k) \right|^2 \right), \quad f = 1, \dots, 80.
    \]
This process produces an 80-dimensional log-Mel spectrogram that effectively captures the time-frequency characteristics of speech.

For the visual stream, we adopt ConvNeXt\cite{woo2023convnext} network to extract robust facial features from each video frame. First, we used  pretrained face detector\cite{schroff2015facenet} to localize the facial region within each frame. The detected face is then cropped, resized to a fixed resolution , and normalized to ensure consistency across varying illumination conditions.

Once pre-processed, the facial image $I \in \mathbb{R}^{H \times W \times 3}$ is fed into the pre-trained ConvNeXt network. ConvNeXt, which embodies modern design principles,  begins with a "patchify" stem that partitions the input image into non-overlapping patches using a $4 \times 4$ convolution with stride 4. Compared to traditional neural networks such as ResNet , this operation effectively reduces the spatial dimensions while preserving local structure. 

Following the stem, a series of convolutional blocks—each incorporating large kernel sizes (typically $7 \times 7$ depthwise convolutions) and layer normalization instead of batch normalization—process the input to generate a high-dimensional feature map. Let the output be denoted as $F \in \mathbb{R}^{\frac{H}{s} \times \frac{W}{s} \times C}$, where $s$ is the overall downsampling factor and $C$ is the number of channels.

To obtain a compact, global representation of the face, we apply global average pooling over the spatial dimensions:
\[
f = \frac{1}{\frac{H}{s} \cdot \frac{W}{s}} \sum_{i=1}^{\frac{H}{s}} \sum_{j=1}^{\frac{W}{s}} F(i,j),
\]
resulting in a feature vector $f \in \mathbb{R}^{C}$. This formulation not only preserves the essential texture and structural information but also provides translation invariance—critical for capturing subtle facial expressions.

In addition to the global view, we generate local views by applying spatio-temporal crops to the video frames. These crops focus on specific temporal segments of the video, allowing us to capture more detailed facial movements or expressions that occur over time. The local views are generated by selecting 1-second temporal segments of the video and performing random augmentations, such as cropping, flipping, or rotating. These enhanced spatio-temporal crops allow the model to focus on particular regions of interest within the video, capturing short-term variations in facial action units that are critical for accurate recognition.

Both the global and local views for audio and video are then projected into a shared embedding space. For the audio, the local views are reshaped into smaller time-frequency patches, while the video local views are reshaped into smaller spatio-temporal cuboids. These views are processed through their respective encoders, and the outputs are flattened into vectors, which are then projected onto the embedding space. These embeddings are subsequently fed into the cross-modal alignment and fusion stage, where complementary information from the audio and visual streams will be integrated for  facial action unit recognition.

\subsection{Advanced Cross-Modal Fusion}
In this work, we introduce a cross-modal alignment and fusion module that effectively synchronizes audio and visual features to enhance the accuracy of AU recognition\cite{lee2023modality}. This module leverages multi-scale self-attention mechanisms and adaptive sliding windows to capture temporal dependencies and fuse complementary information across modalities.

The multi-scale self-attention mechanism operates at different temporal resolutions, enabling the model to focus on both fine-grained and long-term patterns in the audio and video streams. By employing adaptive sliding windows, we account for varying temporal contexts, ensuring that the fusion process dynamically adapts to different events in the sequence. This approach allows the model to capture subtle variations in facial expressions while maintaining a coherent understanding of the overall context, thus improving robustness.

The fusion process can be mathematically expressed as follows:

\[
F_{\text{fusion}}(t) = \sum_{i=1}^{N} \left( \alpha_i \cdot A_i(t) + \beta_i \cdot V_i(t) \right)
\]

Where  $F_{\text{fusion}}(t)$ represents the fused feature at time $t$.  $A_i(t)$ and $V_i(t)$ are the audio and visual features at scale $i$. $\alpha_i$ and $\beta_i$ are the learned weights that control the importance of each modality at scale $i$. $N$ is the number of scales considered in the multi-scale attention process.

The fusion strategy ensures that both audio and visual features contribute meaningfully, with the weights adjusted dynamically based on the context, allowing the model to focus on the most relevant aspects of the input data.

\subsection{Temporal Modeling \& Classification Optimization}

Temporal Modeling is implemented using a Temporal Convolutional Network (TCN)\cite{lea2017temporal} to capture sequential patterns in our audio-visual features. We input the high-dimensional embeddings from the cross-modal fusion module into a TCN that employs several layers of 1D convolutions with increasing dilation factors. This design captures both local and long-range dependencies while ensuring causality, as each convolution uses only current and past inputs. By directly leveraging TCN for temporal modeling, we bypass the limitations of traditional recurrent networks and simplify the overall architecture. Residual connections further help maintain information flow and ease training.

Following temporal feature extraction, the output is fed into a multilayer perceptron (MLP) \cite{popescu2009multilayer}classifier. The MLP consists of two fully connected layers: the first with 512 units and ReLU activation, followed by dropout to mitigate overfitting, and the final layer applies softmax activation to produce action unit predictions. We optimize the entire network using cross-entropy loss on the Aff-Wild2\cite{kollias2019deep,kollias2019expression,kollias2020analysing,kollias2019face,kollias2021affect,kollias2021analysing,kollias2021distribution,kollias2024distribution,kolliasadvancements,zafeiriou2017aff} dataset. This approach provides a streamlined yet effective framework for modeling temporal dynamics and enhancing the accuracy and robustness of AU recognition in complex video data.

\section{Experiments}

\subsection{Dataset}

In this study, we use the Aff-Wild2\cite{kollias2019deep,kollias2019expression,kollias2020analysing,kollias2019face,kollias2021affect,kollias2021analysing,kollias2021distribution,kollias2024distribution,kolliasadvancements,zafeiriou2017aff} dataset, a large-scale extension of the original Aff-Wild dataset designed for emotion recognition and action unit (AU) detection in spontaneous facial expressions. Aff-Wild2 consists of 564 videos, totaling approximately 2.8 million frames, with annotations for seven basic expressions, twelve AUs, and continuous values for valence and arousal. 
The dataset includes 554 participants from diverse demographic backgrounds, including variations in age, ethnicity, and nationality, representing a wide range of emotional and environmental contexts. The annotations are frame-level, making it unique in that it provides both audio and video annotations, enabling cross-modal emotion recognition. Aff-Wild2 is also notable for its real-world complexity, capturing spontaneous emotional expressions in natural settings, which is crucial for training models that generalize well in practical applications.

Overall, the Aff-Wild2 dataset provides a comprehensive resource for emotion recognition and AU detection tasks, making it an invaluable asset for advancing research in these fields.

\subsection{Results on Validation set}

For the evaluation of AU recognition, we measure model performance using the F1 score for each action unit. To improve robustness and generalization, we conducted six-fold cross-validation on randomly segmented annotated data. Table~\ref{tab:AU_results} presents the F1 scores for each fold (fold-0 to fold-5). The final row highlights the highest F1 score achieved across all folds in bold, representing the best model performance.The results of the validation set show that our method has achieved excellent performance.

\begin{table}
  \centering
  \caption{Comparison Results of Samples Used for Fusion and Hard Voting Outcomes}
  \label{tab:AU_results}
  \begin{tabular}{ccl}
    \toprule
     & F1 score (\%)\\
    \midrule
    \text{fold-0}& 54.20\\
    \text{fold-1}&  54.94\\
    \text{fold-2}& 55.07 \\
    \text{fold-3}& 55.18\\
    \text{fold-4}& 55.85 \\
    \textbf{fold-5}& \textbf{56.11}\\    
    \bottomrule
  \end{tabular}
\end{table}

\section{Conclusion}

In this study, we proposed an audio-visual fusion framework for facial action unit (AU) detection in naturalistic settings. By leveraging pre-trained Whisper and ConvNeXt models, we effectively capture complementary information from both modalities. A key innovation is the introduction of global and local views, where global views provide holistic context, and local views—enhanced spatio-temporal and time-frequency crops—focus on fine-grained, discriminative features.
To enhance temporal modeling, we integrate a cross-modal alignment module with multi-scale self-attention and adaptive sliding windows, ensuring robust fusion of dynamic features. Additionally, a Temporal Convolutional Network (TCN) captures both short-term and long-range dependencies, improving AU recognition.

Experimental results on the Aff-Wild2 dataset demonstrate strong AU recognition performance, confirming the effectiveness of our method. These findings highlight the impact of global-local feature extraction and advanced cross-modal fusion. Future work will explore refined attention mechanisms and domain adaptation to further improve AU recognition in real-world applications.

{
    \small
    \bibliographystyle{ieeenat_fullname}
    \bibliography{main}
}


\end{document}